\documentclass[a4paper]{article}

\usepackage{INTERSPEECH2020}
\usepackage{amsthm}
\usepackage{amsmath}
\newtheorem{theorem}{Theorem}
\newtheorem{corollary}{Corollary}[theorem]

\title{Statistical Testing on ASR Performance via Blockwise Bootstrap}
\name{Zhe Liu, Fuchun Peng}
\address{
  Facebook AI, Menlo Park, CA, USA}
\email{zheliu@fb.com, fuchunpeng@fb.com}

\begin{document}

\maketitle
\begin{abstract}
A common question being raised in automatic speech recognition (ASR) evaluations is how reliable is an observed word error rate (WER) improvement comparing two ASR systems, where statistical hypothesis testing and confidence interval (CI) can be utilized to tell whether this improvement is real or only due to random chance. The bootstrap resampling method has been popular for such significance analysis which is intuitive and easy to use. However, this method fails in dealing with dependent data, which is prevalent in speech world - for example, ASR performance on utterances from the same speaker could be correlated. In this paper we present blockwise bootstrap approach - by dividing evaluation utterances into nonoverlapping blocks, this method resamples these blocks instead of original data. We show that the resulting variance estimator of absolute WER difference between two ASR systems is consistent under mild conditions. We also demonstrate the validity of blockwise bootstrap method on both synthetic and real-world speech data.
\end{abstract}

\section{Introduction}

The most widely used metric for measuring the performance of an automatic speech recognition (ASR) system is the word error rate (WER), which is derived from the Levenshtein distance \cite{navarro2001guided} working at the word level:
\begin{equation}
    WER=\frac{\sum_{s=1}^n e_s}{\sum_{s=1}^n m_s},
\end{equation}
where $m_s$ is the number of words in the $s$th sentence (i.e. reference text of audio) of the evaluation dataset, and $e_s$ represents the sum of insertion, deletion and substitution errors computed from the dynamic string alignment of the recognized word sequence with the reference word sequence. The WER may also be referred to as the length normalized edit distance \cite{niessen2000evaluation}.

A practical question being commonly raised in ASR system evaluations is that how reliable is an observed improvement of ASR system \emph{B} comparing to ASR system \emph{A}. For example, if we obtained an absolute 0.2\% WER reduction, how could we tell if this improvement is real and not due to random chance.

Here is where the statistical hypothesis testing comes into play. The use of statistical testing in ASR evaluations has been previously explored \cite{gillick1989some, pallett1993benchmark, strik2000comparing, bisani2004bootstrap, vilar2007human, vilar2008efficient}. In particular, the work of \cite{bisani2004bootstrap} presents a \emph{bootstrap} method for significance analysis on ASR evaluations which makes no distributional approximations and the results are immediately interpretable in terms of WER.

To be more specific, suppose we have a sequence of independent and identically distributed (i.i.d.) random variable $\{Z_s\}_{s=1}^n$ and we are interested in estimating the variance of some statistic $T(Z_1, ..., Z_n)$. The bootstrap method \cite{efron1994introduction, efron2003second} resamples data from the empirical distribution of $\{Z_s\}_{s=1}^n$, and then recalculates the statistic $T$ on each of these ``bootstrap'' samples. Then the variance of $T(Z_1, ..., Z_n)$ can be estimated from the sample variance of these computed statistics.

For the ASR systems comparison problem that we raised previously, authors in \cite{bisani2004bootstrap} proposed the idea of using bootstrap approach above to resample (with replacement) the utterances in the evaluation dataset for each replicate, and then estimate the probability that the absolute WER difference of ASR system \emph{B} versus system \emph{A}
\begin{equation}
\label{diff}
    \Delta W:={WER}^B - {WER}^A=\frac{\sum_{s=1}^n (e_s^B - e_s^A)}{\sum_{s=1}^n m_s}
\end{equation}
is positive, where ASR systems \emph{A} and \emph{B} have word error counts $e_s^A$ and $e_s^B$ on the $s$th sentence, respectively. Notice that we should calculate the difference in the number of errors of the two systems on identical bootstrap samples.

However, one of the key issue confronting bootstrap resampling approximations is how to deal with dependent data \cite{kreiss2011bootstrap}. This is particularly the case for speech data since the speech recognition errors could be highly correlated across different utterances if they are 1) from the same speaker, 2) similar in the uttered sentence (e.g. in the same domain or topic), or both. When the dependent structure across $\{Z_s\}_{s=1}^n$ is nontrivial, the true sampling distribution of $T(Z_1, ..., Z_n)$ would depend on the joint distribution of $\{Z_s\}_{s=1}^n$ and thus the bootstrap samples should preserve such dependent structure as well. Unfortunately, the reshuffled samples obtained from the ordinary bootstrap method break such dependence, and thus lead to wrong variance estimations of the statistic. In particular, ASR errors on dependent utterances could always be positively correlated, which typically makes confidence intervals computed by bootstrap much narrower than what they should be. This might lead to over-optimistic conclusions due to false-positive discovery.

In this paper, we present the \emph{blockwise bootstrap} approach for statistical testing of ASR performance. By dividing evaluation utterances into nonoverlapping blocks, we resamples these blocks instead of original data points (i.e. word errors) of utterances. The idea of blockwise bootstrap was initially developed in the work of \cite{hall1985resampling, carlstein1986use} for dealing with dependent time series data. Since then, there has been a line of research in statistical literature on various block construction methods and theoretical comparisons of them \cite{liu1992moving, politis1992circular, politis1994stationary, morvai1996nonparametric, lahiri1999theoretical, nordman2009note}. To the best of our knowledge, our work is the ﬁrst to introduce the blockwise bootstrap approach to address dependent speech data in ASR performance evaluations and illustrate how it helps calculate valid confidence intervals for absolute WER difference.

The rest of the paper is organized as follows. Section \ref{methodology} introduces the use of blockwise bootstrap on ASR evaluation problem. Section \ref{properties} shows the statistical property that the blockwise variance estimator is consistent under mild conditions. Section \ref{simulation} and Section \ref{real} demonstrate the validity of blockwise bootstrap method on simulated synthetic data and real-world speech data. We conclude in Section \ref{conclusion}.

\section{Methods}
\label{methodology}
In this section, we describe the blockwise bootstrap method to compute the confidence interval of the absolute WER difference $\Delta W$ as in the formula (\ref{diff}).

Given two ASR systems \emph{A} and \emph{B} in the comparison, assume we have their evaluation results on $n$ utterances as follows:
\begin{equation}
    (m_1, e_1^A, e_1^B), (m_2, e_2^A, e_2^B), \ldots, (m_n, e_n^A, e_n^B)
\end{equation}
where for any $s=1,\ldots,n$, let $m_s$ be the number of words in the reference of $s$th utterance, $e_s^A$ and $e_s^B$ represent the numbers of word errors in ASR systems \emph{A} and \emph{B}, respectively. The statistic that we are interested in is the absolute WER difference $\Delta W$ comparing system \emph{B} versus system \emph{A}.

Suppose the evaluation results data above can be partitioned into $K$ nonoverlapping blocks (e.g. by speakers) such that word error counts within each block are correlated while the dependency of word error counts from different blocks are negligible
\begin{equation}
\{(m_s, e_s^A, e_s^B)\}_{s\in S_k}, k=1,\ldots,K
\end{equation}
where $\cup_k S_k=\{1,2,\ldots,n\}$, and $S_i\cap S_j=\emptyset$ for any $i$, $j$. 
Then the blockwise bootstrap method works as follows.

For any $b=1,\ldots,B$ where $B$ is a large number, we randomly sample (with replacement) $K$ elements $\{S_{k'}^{(b)}\}_{k'=1,\ldots,K}$ from the set $\{S_k\}_{k=1,\ldots,K}$ to generate a bootstrap sample of evaluation results data
\begin{equation}
\{(m_s, e_s^A, e_s^B)\}_{s\in S_{k'}^{(b)}}, k'=1,\ldots,K
\end{equation}
where each $S_{k'}^{(b)}\in\{S_k\}_{k=1,\ldots,K}$. Then for this bootstrap replicate sample, the statistic is computed as
\begin{equation}
\Delta W^{(b)}=\frac{\sum_{k'=1}^K\sum_{s\in S_{k'}^{(b)}}(e_s^B-e_s^A)}{\sum_{k'=1}^K\sum_{s\in S_{k'}^{(b)}}m_s}.
\end{equation}
Once we have all $\{\Delta W^{(b)}\}_{b=1,\ldots,B}$, then the 95\% confidence interval for $\Delta W$ can be determined by the empirical percentiles at 2.5\% and 97.5\% of the bootstrap sample statistics
\begin{equation}
\label{per}
(\Delta W_{2.5\%}^{\emph{blockBoot}}, \Delta W_{97.5\%}^{\emph{blockBoot}}).
\end{equation}

Alternatively, the uncertainty of $\Delta W$ can be quantified by its standard error, which can be approximated by the sample standard deviation of the $B$ bootstrap sample statistics
\begin{equation}
se^{\emph{blockBoot}}(\Delta W) = \sqrt{\frac{\sum_{b=1}^B(\Delta W^{(b)}-m^{\emph{blockBoot}}(\Delta W))^2}{B-1}}
\end{equation}
where $m^{\emph{blockBoot}}(\Delta W)=\frac{1}{B}\sum_{b=1}^B \Delta W^{(b)}$. Then based on the Gaussian approximation, the 95\% confidence interval for $\Delta W$ can be obtained by
\begin{equation}
\label{se}
(m^{\emph{blockBoot}}(\Delta W)\pm1.96\cdot se^{\emph{blockBoot}}(\Delta W)).
\end{equation}

Generally speaking, the percentile confidence intervals (\ref{per}) always give similar results with Gaussian approximation confidence intervals (\ref{se}) when $B$ is large, unless the corresponding bootstrap sample statistics are highly skewed, in which case the former ones are preferred.

Note that in this paper we mainly focus on the confidence intervals of absolute WER difference. Similarly, the blockwise bootstrap method can also be utilized to compute the confidence intervals for the WER itself as well as relative WER difference between two ASR systems.

\section{Theoretical Properties}
\label{properties}
We work out some statistical theories to show that the blockwise variance estimator of $\Delta W$ is consistent under mild conditions. That is, as the number of evaluation data points increases indefinitely, the resulting sequence of variance estimates converges to the truth variance \cite{amemiya1985advanced}.

For simplicity, we assume all utterances in the evaluation dataset have the same number of words, that is, $m_i=m$ for all $i=1,\ldots,n$. Let's denote $Z_i:=(e_i^B-e_i^A)/m$. Then we have the statistic of interest written as
\begin{equation}
    \Delta W_n = \frac{1}{n}\sum_{i=1}^n Z_i
\end{equation}
where the subscript $n$ in $\Delta W_n$ indicates the number of samples (i.e. utterances) corresponding to the quantity. Further, by dividing the $n$ utterances into nonoverlapping blocks, suppose each block has same number of utterances, denoted as $d_n$. Let $K_n=[n/d_n]$ as the number of blocks. Note that the assumptions of having same number of words in each utterance as well as same number of utterances in each block are for the sake of simplicity, the results below still hold if the equality assumptions here are relaxed to be in the same order of $n$. 

Without the loss of generality, assume that the blocks are consecutive and thus the $k$th block consists of the sequence $Z_{(k-1)d_n +1}, \ldots,Z_{kd_n}$ where $k=1,\ldots,K_n$. We further let
\begin{equation}
    \Delta W_{d_n}^{(k-1)d_n+1}:=\frac{1}{d_n}\sum_{i=(k-1)d_n+1}^{kd_n}Z_i
\end{equation}
where the subscript $d_n$ in $\Delta W_{d_n}^{(k-1)d_n+1}$ for the $k$th block represents the block size and the superscript $(k-1)d_n+1$ indicates the index of starting variable in the block. Notice that $\frac{1}{K_n}\sum_{k=1}^{K_n}\Delta W_{d_n}^{(k-1)d_n+1}=\Delta W_n$.

Consider the blockwise variance estimator (i.e. standardized sample variance of blockwise estimators of the statistic)
\begin{equation}
    \hat\sigma_n^2:=\frac{d_n}{K_n}\sum_{k=1}^{K_n}{\left(\Delta W_{d_n}^{(k-1)d_n+1}-\Delta W_n\right)}^2.
\end{equation}
The following theorem establishes its $L_2$-consistency result.

\begin{theorem}
\label{thm:main}
Assume the asymptotic variance of $\Delta W_n$ is 
\begin{equation}
    \lim_{n\rightarrow\infty}n\mathbf{E}(\Delta W_n-\mathbf{E}(\Delta W_n))^2=\sigma^2\in(0,\infty)
\end{equation}
and $\mu=\mathbf{E}(Z_i)$ for any $i=1,\ldots,n$. Let $d_n$ be s.t. $d_n\rightarrow\infty$ and $K_n\rightarrow\infty$ as $n\rightarrow\infty$. If $n^2\mathbf{E}(\Delta W_n-\mu)^4$ is uniformly bounded, and for any $1\leq k< k'\leq K_n$, the sequence of $Z_{(k-1)d_n +1}, \ldots,Z_{kd_n}$ and the sequence of $Z_{(k'-1)d_n +1}, \ldots,Z_{k'd_n}$ are uncorrelated, then
\begin{equation}
    \hat\sigma_n^2\rightarrow_{L_2}\sigma^2\;as\;n\rightarrow\infty.
\end{equation}
\end{theorem}

The proof of the theorem is deferred to Appendix (accompanied by additional files). According to Theorem \ref{thm:main}, we require both the number of blocks $K_n$ and number of utterances in each block $d_n$ go to infinity as the number of utterances $n$ grows to infinity. This is reasonable for speech evaluation data collection if the blocks are partitioned by different speakers or topics.

On the other hand, the uncorrelated assumption between different blocks seems strong, because in practice it's possible that word errors over different blocks are also weakly correlated. We relax this assumption in the corollary below.

\begin{corollary}
\label{thm:corollary}
Theorem \ref{thm:main} still holds if the assumption of uncorrelated blockwise variables is relaxed as follows: for any $\epsilon>0$, if $n$ is large enough, for any $1\leq k<k'\leq K_n$ and $(k-1)d_n + 1\leq i<j\leq kd_n$, $(k'-1)d_n + 1 \leq i'<j'\leq k'd_n$ assume
\begin{equation}
    \mathbf{E}(|(Z_i-\mu)(Z_j-\mu)(Z_{i'}-\mu)(Z_{j'}-\mu)|)\leq \epsilon.
\end{equation}
\end{corollary}

The proof of the corollary is also deferred to Appendix (accompanied by additional files).

\section{Simulation Experiments}
\label{simulation}
In this section, we conduct simulation experiments to show that the blockwise bootstrap approach is capable to generate valid confidence intervals for absolution WER differences between two ASR systems and is superior to the ordinary bootstrap when the utterances in the evaluation dataset are dependent.

\subsection{Setup}
In this simulation experiments, we generate synthetic data, i.e. counts of ASR errors, to measure the performance of ordinary bootstrap and blockwise bootstrap methods. We assume the total number of utterances in the evaluation set is $n=3$,$000$ and number of words in each utterance is equally $m=100$. For the two ASR systems \emph{A} and \emph{B} in the comparison, suppose the ``ground-truth'' WERs are given by $WER^A=10.0\%$ and $WER^B=9.5\%$ respectively. Thus the absolution WER difference $\Delta W$ between them is $-0.5\%$.

Under the scenario that the numbers of errors from different utterances are independent with each other, we generate the number of errors for each utterance from the binomial distribution $Binom(m=100,p)$ where $p=WER^A$ or $WER^B$ depending on which ASR system was used. On the other hand, when the ASR errors are dependent across different utterances, we need to make additional correlation structure assumption while keeping the marginal distribution of error count on each utterance to be binomial distributed.

Here we assume the numbers of errors across different utterances are block-correlated, that is, for any two utterances, their ASR errors are correlated if they belong to the same block while their errors are independent if they belong to different blocks. Without the loss of generality, suppose the blocks are consecutive and the size of block (i.e. number of utterances in each block) is denoted as $d$. Follow the steps below to generate ASR errors for each block and each ASR system:
\begin{enumerate}
    \item Generate a sample $(v_1,\ldots,v_d)$ from multivariate Gaussian distribution $N(0, \Sigma_d)$, where $\Sigma_d$ is an $d$-by-$d$ covariance matrix with $\sigma_{ij}=1$ if $i=j$ and $\sigma_{ij}=\rho$ if $i\not=j$. Here $\sigma_{ij}$ is the $(i,j)$-th element of $\Sigma_d$;
    \item Turn $(v_1,\ldots,v_d)$ into correlated uniforms $(u_1,\ldots,u_d)$ where $u_s=\Phi(v_s)$ for $s=1,\ldots,d$ and $\Phi(\cdot)$ is the Gaussian cumulative distribution function (CDF);
    \item Generate correlated Binomial samples $(e_1,\ldots,e_d)$ by inverting the Binomial CDF: $e_s=Q_{binom}(u_s, m, p)$ for $s=1,\ldots,d$, where $Q_{binom}(\cdot,m,p)$ is the inverse of the Binomial CDF.
\end{enumerate}
The counts of word errors for different blocks and different ASR systems are generated independently. For both ordinary bootstrap and blockwise bootstrap methods in the comparison, we set the resampling size $B=1$,$000$. The block size $d$ and correlation parameter $\rho$ are varied in our experiment.

While in this and next section we mainly focus on the comparison again ordinary bootstrap method, there also exists other parametric approaches (e.g. in \cite{vilar2008efficient}) without the need of Monte Carlo resampling. They typically give similar results with bootstrap but are less preferred than bootstrap especially when these distributional assumptions are violated in practice.

\begin{figure}
  \begin{minipage}[c]{0.494\linewidth}
   \centering
    \includegraphics[width=\linewidth]{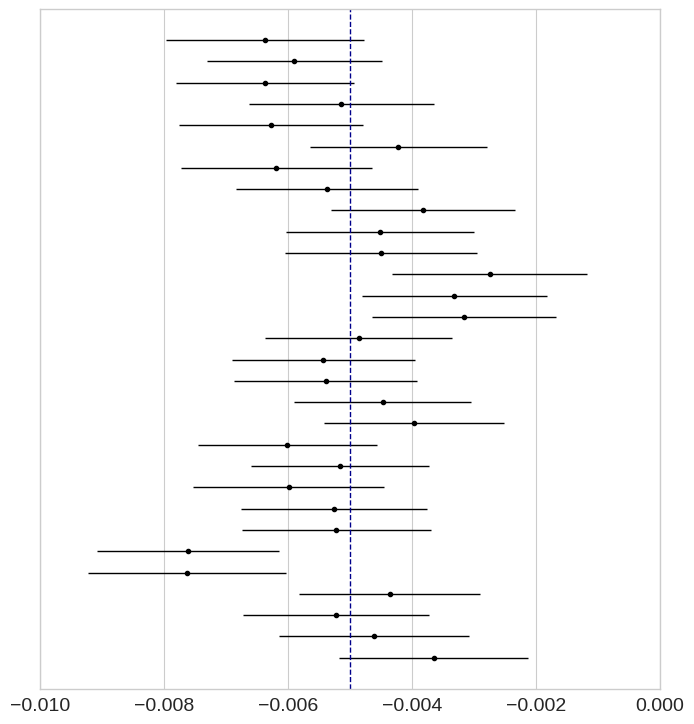} \\
    (a) Bootstrap
    \end{minipage}
  \hfill
  \begin{minipage}[c]{0.494\linewidth}
    \centering
    \includegraphics[width=\linewidth]{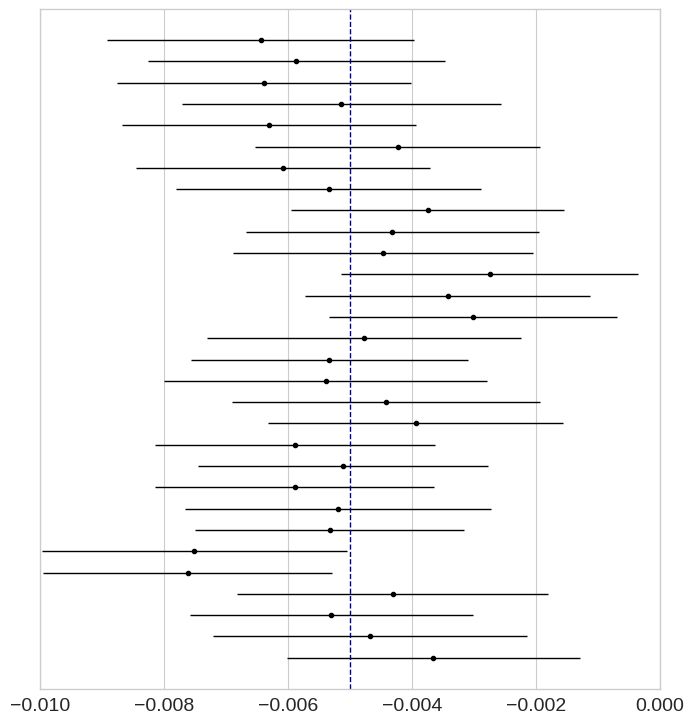} \\
    (b) Blockwise Bootstrap
  \end{minipage}
  \caption{Visualisation of confidence intervals computed on the first 30 simulated data ($d=5, \rho=0.4$), from both ordinary bootstrap and blockwise bootstrap. The vertical blue line represents the true absolute WER difference $\Delta W=-0.5\%$.}
  \label{fig:simulation}
\end{figure}

\subsection{Results}
Strictly speaking, a 95\% confidence interval means that if we were able to have 100 different datasets from the same distribution of the original data and compute a 95\% confidence interval based on each of these datasets, then approximately 95 of these 100 confidence intervals will contain the true value of the statistic of interest \cite{neyman1937x, stuart1963advanced, cox1979theoretical}. In our experiments, for each setup of various block size and correlation parameter, we replicate the simulation for 1,000 times. Therefore if confidence intervals were computed correctly, then approximately 950 of these confidence intervals would contain the true absolute WER difference $\Delta W=-0.5\%$.

Seen from Table~\ref{tab:simulation}, the blockwise bootstrap method always gives valid confidence intervals since the percentage of confidence intervals that contain the true $\Delta W$ is very close to 95\%, regardless of block size ($d=5$ or $30$) or correlation parameter ($\rho=0, 0.05, 0.1, 0.2$ or $0.4$). As the block size or correlation increases, the width of confidence intervals becomes larger. On the other hand, the ordinary bootstrap method fails to generate correct confidence intervals when the data is dependent ($\rho>0$), since its percentage of confidence intervals that contain the true $\Delta W$ is much lower than 95\%.

Figure~\ref{fig:simulation} plots the confidence intervals computed on the first 30 simulated data for $d=5$ and $\rho=0.4$, where we can see that the confidence intervals from the blockwise bootstrap method are wider and capture more true values of $\Delta W=-0.5\%$.

It is worth noting that when the dependency across different utterances is relatively strong, the width of confidence intervals generated by ordinary bootstrap is less than the half of the valid ones generated by blockwise bootstrap, in which case ordinary bootstrap might lead to over-optimistic discovery.

\begin{table}
 \caption{Comparison results of ordinary bootstrap and blockwise bootstrap methods on simulated data with various block size ($d$) and correlation parameter ($\rho$), where the average width of confidence intervals and the percentage of times that confidence intervals contain the true $\Delta W$ are shown.}
  \centering
  \resizebox{\columnwidth}{!}{%
  \begin{tabular}{cl|cc|cc}
    \toprule
    & & \multicolumn{2}{c}{\bf{Bootstrap}} & \multicolumn{2}{|c}{\bf{Blockwise Bootstrap}} \\
    \cmidrule(r){3-4}
    \cmidrule(r){5-6}                                   
    \shortstack{\emph{Block Size} \\ ($d$)} & \shortstack{\emph{Correlation} \\ ($\rho$)} & \emph{Width} & \shortstack{\emph{\% contains} \\ \emph{the truth}} & \emph{Width} & \shortstack{\emph{\% contains} \\ \emph{the truth}} \\
    \midrule
    $d=5$ & $\rho=0$ & 0.0030 & 94.1\% & 0.0030 & 94.7\%  \\
    & $\rho=0.05$ & 0.0030 & 92.7\% & 0.0033 & 95.2\% \\
    & $\rho=0.1$ & 0.0030 & 90.1\% & 0.0035 & 94.3\% \\
    & $\rho=0.2$ & 0.0030 & 86.2\% & 0.0040 & 94.9\% \\
    & $\rho=0.4$ & 0.0030 & 76.9\% & 0.0048 & 94.0\% \\       
    \midrule
    $d=30$ & $\rho=0$ & 0.0030 & 94.1\% & 0.0030 & 94.7\%  \\
    & $\rho=0.05$ & 0.0030 & 78.1\% & 0.0046 & 95.2\% \\
    & $\rho=0.1$ & 0.0030 & 69.2\% & 0.0058 & 94.9\% \\
    & $\rho=0.2$ & 0.0030 & 54.4\% & 0.0077 & 94.7\% \\
    & $\rho=0.4$ & 0.0030 & 41.2\% & 0.0105 & 95.9\% \\    
    \bottomrule
  \end{tabular}
  }
  \label{tab:simulation}
\end{table}

\section{Real Data Experiments}
\label{real}
In this section, we also apply the blockwise bootstrap approach on two real-world speech datasets and demonstrate how it helps compute the confidence intervals of absolute WER difference between two ASR systems. We consider the following two ASR evaluation datasets in this analysis
\begin{itemize}
\item \emph{Conversational Speech} data. This dataset was collected through crowd-sourcing from a data supplier for ASR, and the data was properly anonymized. It consists of 235 conversions with more than 20 topics that are common in daily life, including family, travel, foods, etc;
\item \emph{Augmented Multi-Party Interaction (AMI) Meeting} data. The dataset \cite{carletta2005ami, aran2010multimodal} includes scenario meetings (with roles assigned for participants) and non-scenario meetings (where participants were free to choose topics). For scenario meetings, each session is divided into 4 one-hour meetings. Each meeting has 4 participants.
\end{itemize}
Since our purpose is to evaluate ASR performance, we only use the ``dev'' and ``eval'' splits of the entire datasets above.

We use in-house developed conversation ASR system in this investigation: a baseline model (denoted as system \emph{A}) and an improved model (denoted as system \emph{B}), and we are interested in computing a 95\% confidence interval of the absolute WER difference between the two ASRs. If the upper bound of such confidence interval is negative, then we can tell that this improvement is real and not due to random chance.

To apply the blockwise bootstrap method, we need to define the correlated block structures among the utterances in the evaluation data (merged ``dev'' and ``eval'' splits). For Conversation data, it's natural to treat each conversation as a single separated block since the same topics were being discussed. For AMI Meeting dataset, we treat the utterances from each speaker in each (either scenario or non-scenario) meeting as a block. By doing that, for any two utterances, we assume their ASR errors are correlated if they belong to the same block while the errors have very weak correlations if they belong to different blocks. Table~\ref{tab:data} shows details of the two evaluation datasets in terms of number of utterances, number of total words, and number of  correlated blocks.

We apply both ordinary bootstrap and blockwise bootstrap methods on the two evaluation datasets. Results are shown in Table \ref{tab:real}. Again, we observe that the confidence intervals computed from blockwise bootstrap are much wider than the ones generated from ordinary bootstrap: around 1.5 times wider on Conversation data and 2 times wider on AMI Meeting data. 
Also, we can see that confidence intervals computed from the empirical percentiles at 2.5\% and 97.5\% of bootstrap samples are almost the same with the ones computed from Gaussian approximation.

\begin{figure}
  \begin{minipage}[c]{0.495\linewidth}
    \centering
    \includegraphics[width=\linewidth]{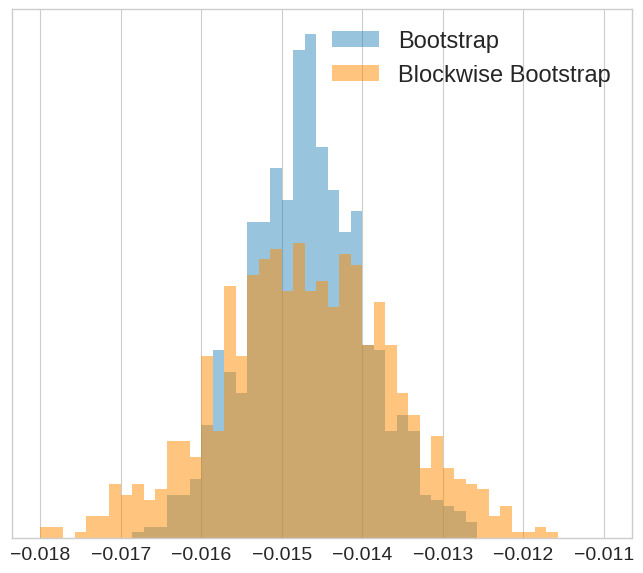} \\
    (a) \emph{Conversation}
    \end{minipage}
  \hfill
  \begin{minipage}[c]{0.495\linewidth}
    \centering
    \includegraphics[width=\linewidth]{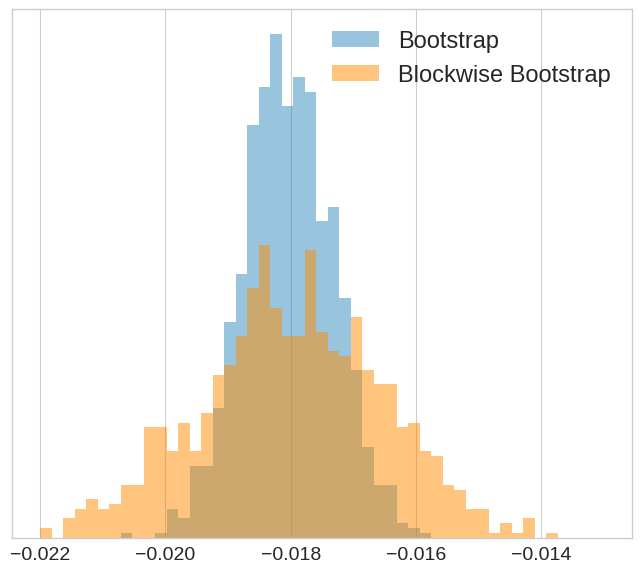} \\
    (b) \emph{AMI Meeting}
  \end{minipage}
  \caption{Histograms of absolute WER difference calculated from the bootstrap samples of both bootstrap and blockwise bootstrap methods.}
  \label{fig:real}
\end{figure}

Figure~\ref{fig:real} displays the histograms of absolute WER difference computed from the bootstrap samples, where we can see again that the data distribution from the blockwise bootstrap method is more spread out. Thus confidence intervals computed by ordinary bootstrap  underestimate the standard errors and are much narrower than what they should be, which could lead to over-optimistic conclusions due to false-positive discovery.

\begin{table}
 \caption{Summary of the Conversation Speech and AMI Meeting datasets in the experiments of real data analysis.}
  \centering
  \resizebox{\columnwidth}{!}{%
  \begin{tabular}{l|r|r}
    \toprule
    & \multicolumn{2}{|c}{\bf{Evaluation Dataset}} \\
    \cmidrule(r){2-3}    
    \emph{Feature}& \emph{Conversation} & \emph{AMI Meeting}  \\
    \midrule
    Number of Utterances & 13,987 & 25,741 \\
    Number of Words & 160,338 & 189,590 \\
    Number of Correlated Blocks & 235 & 135 \\
    \bottomrule
  \end{tabular}
  }
  \label{tab:data}
\end{table}

\begin{table}
 \caption{Results of bootstrap and blockwise bootstrap methods on real-world Conversation and AMI meeting datasets.}
  \centering
  \resizebox{\columnwidth}{!}{%
  \begin{tabular}{cc|c|c}
    \toprule
    & & \multicolumn{2}{|c}{\bf{Evaluation Dataset}} \\
    \cmidrule(r){3-4}    
    \emph{Method} & \emph{Metric}& \emph{Conversation} & \emph{AMI Meeting}  \\
    \midrule
    \bf{Bootstrap} & $\Delta W$ & $-1.47\%$ & $-1.80\%$ \\
    & $se^{\emph{boot}}(\Delta W)$ & $0.074\%$ & $0.067\%$ \\
    & \emph{Percentile CI} & $(-1.61\%, -1.32\%)$ & $(-1.94\%, -1.67\%)$\\
    & \emph{Gaussian Approx. CI} & $(-1.62\%, -1.33\%)$ & $(-1.94\%, -1.67\%)$\\
    \midrule
    \bf{Blockwise} & $\Delta W$ & $-1.47\%$ & $-1.80\%$ \\
    \bf{Bootstrap} & $se^{\emph{blockBoot}}(\Delta W)$ & $0.116\%$ & $0.153\%$ \\
    & \emph{Percentile CI} & $(-1.69\%, -1.24\%)$ & $(-2.09\%, -1.51\%)$\\
    & \emph{Gaussian Approx. CI} & $(-1.69\%, -1.24\%)$ & $(-2.10\%, -1.50\%)$\\
    \bottomrule
  \end{tabular}
  }
  \label{tab:real}
\end{table}

\section{Conclusion}
\label{conclusion}
In this paper, we present blockwise bootstrap approach for statistical testing of ASR performance - by dividing the evaluation utterances into nonoverlapping blocks, this method resamples these blocks instead of original data. We show that the resulting variance estimator of the absolute WER difference is consistent under mild conditions. We also illustrate the validity of blockwise bootstrap method on synthetic and real-world speech data.

Future work might include how to infer the correlated block structures from data, for example, estimating a blockwise sparse correlation matrix across evaluation utterances based on the embeddings of speakers and text sentences.

\pagebreak
\bibliographystyle{IEEEtran}

\bibliography{mybib}

\clearpage
\section{Appendix}
\label{appen}
\subsection{Proof of Theorem \ref{thm:main}}
\begin{proof}
Let's denote 
\begin{equation}
    t_{d_n}^{(k-1)d_n+1}:=\sqrt{d_n}(\Delta W_{d_n}^{(k-1)d_n+1}-\mu).
\end{equation}
Then the variance estimator can be written as
\begin{equation}
\label{thm:estimator}
    \hat\sigma_n^2=\frac{1}{K_n}\sum_{k=1}^{K_n}\left(t_{d_n}^{(k-1)d_n+1}\right)^2-(t_n)^2
\end{equation}
where
\begin{equation}
    t_n=\sum_{k=1}^{K_n}t_{d_n}^{(k-1)d_n+1}/K_n.
\end{equation}
Note that $\mathbf{E}\left(t_{d_n}^{(k-1)d_n+1}\right)=\mathbf{E}(t_n)=0$.

We will first show that the first term of the right hand side of (\ref{thm:estimator}) converges to $\sigma^2$ in $L_2$. Notice that
\begin{align}
    \mathbf{E}\left(t_{d_n}^{(k-1)d_n+1}\right)^2&=d_n\mathbf{E}\left(\Delta W_{d_n}^{(k-1)d_n+1}-\mu\right)^2 \\
    &\rightarrow\sigma^2
\end{align}
as $n\rightarrow\infty$ and (thus) $d_n\rightarrow\infty$. Then it suffices to show
\begin{equation}
    \textbf{Var}\left(\frac{1}{K_n}\sum_{k=1}^{K_n}\left(t_{d_n}^{(k-1)d_n+1}\right)^2\right)\rightarrow 0.
\end{equation}
This is true since
\begin{align}
\label{thm:relax}
    \textbf{Var}\left(\sum_{k=1}^{K_n}\left(t_{d_n}^{(k-1)d_n+1}\right)^2\right)&\leq\sum_{k=1}^{K_n}\mathbf{E}\left(t_{d_n}^{(k-1)d_n+1)}\right)^4 \\
    &\leq K_n C
\end{align}
when $n$ and $d_n$ are sufficiently large. Here the first less than or equal to sign follows from the assumption that for any two blocks $k< k'$, $\Delta W_{d_n}^{(k-1)d_n+1}$ and $\Delta W_{d_n}^{(k'-1)d_n+1}$ are uncorrelated, and the second less than or equal to sign follows from the assumption that $n^2\mathbf{E}(\Delta W_n-\mu)^4$ is uniformly bounded.

Now we only need to show that the second term of the right hand side of (\ref{thm:estimator}) converges to 0 in $L_2$, or equivalently, $t_n$ converges to 0 in $L_4$.

Note that 
\begin{equation}
    t_n=\sqrt{d_n}(\Delta W_n-\mu)
\end{equation}
and thus
\begin{equation}
    \mathbf{E}(t_n^4)=d_n^2\mathbf{E}(\Delta W_n-\mu)^4\rightarrow 0
\end{equation}
as $n\rightarrow \infty$ since $n^2\mathbf{E}(\Delta W_n-\mu)^4$ is bounded.
\end{proof}

\subsection{Proof of Corollary \ref{thm:corollary}}
\begin{proof}
It suffices to show
\begin{equation}
    \textbf{Var}\left(\frac{1}{K_n}\sum_{k=1}^{K_n}\left(t_{d_n}^{(k-1)d_n+1}\right)^2\right)\rightarrow 0
\end{equation}
under the relaxed assumption. Consider for any $k<k'$
\begin{align}
    &\;\textbf{Cov}\left(\left(t_{d_n}^{(k-1)d_n+1}\right)^2, \left(t_{d_n}^{(k'-1)d_n+1}\right)^2\right) \\
    &\leq\mathbf{E}\left(t_{d_n}^{(k-1)d_n+1}t_{d_n}^{(k'-1)d_n+1}\right)^2 \\
    &=d_n^2\mathbf{E}\left(\Delta W_{d_n}^{(k-1)d_n+1}-\mu\right)^2\left(\Delta W_{d_n}^{(k'-1)d_n+1}-\mu\right)^2 \\
    &=\frac{1}{d_n^2}\mathbf{E}\left(\sum_{\substack{i=(k-1)d_n\\+1}}^{kd_n}(Z_i-\mu)\right)^2\left(\sum_{\substack{i'=(k'-1)d_n\\+1}}^{kd_n}(Z_{i'}-\mu)\right)^2.
\end{align}
Then under the assumption that $\mathbf{E}(|(Z_i-\mu)(Z_j-\mu)(Z_{i'}-\mu)(Z_{j'}-\mu)|)\leq \epsilon$ if $n$ is sufficiently large for any $i,j$ and $i',j'$, we have
\begin{align}
    &\textbf{Var}\left(\frac{1}{K_n}\sum_{k=1}^{K_n}\left(t_{d_n}^{(k-1)d_n+1}\right)^2\right) \\
    &\leq \frac{1}{K_n^2}\sum_{k=1}^{K_n}\mathbf{E}\left(t_{d_n}^{(k-1)d_n+1)}\right)^4\\
    &\quad\;+\frac{2}{K_n^2}\sum_{1\leq k<k'\leq K_n}\mathbf{E}\left(t_{d_n}^{(k-1)d_n+1}t_{d_n}^{(k'-1)d_n+1}\right)^2 \\
    &\leq\frac{1}{K_n^2}(K_n C+K_n^2\epsilon)=\frac{C}{K_n}+\epsilon.
\end{align}
which converges to 0 as $n\rightarrow\infty$.
\end{proof}

\end{document}